\documentclass[preprint]{acmsiggraph}  %

\usepackage[utf8]{inputenc}
\usepackage[T1]{fontenc}

\TOGonlineid{121}
\TOGvolume{$ $}
\TOGnumber{SIGGRAPH Asia 2016}
\TOGarticleDOI{1111111.2222222}
\usepackage[table,usenames,dvipsnames]{xcolor}
\usepackage{amsmath}
\usepackage{amsfonts}
\usepackage{amssymb}
\usepackage{microtype}
\usepackage{color}
\usepackage{textcomp}
\usepackage{booktabs}
\usepackage[capitalise]{cleveref}
\usepackage{textcomp}
\usepackage{titlesec}
\titlespacing*\paragraph{0pt}{0pt}{1em}
\usepackage[scaled=0.85]{beramono}

\ifdefined\useTikZ

	\usepackage{tikz,pgfplots}
	\pgfplotsset{compat=1.3}
	\usetikzlibrary{backgrounds,external,calc,arrows,positioning}

	\tikzset{every picture/.style={tight background,>=stealth,font=\sffamily}}

	\newcommand{\inputTikZfigure}[1]{\input{figures/#1}}

\else %

	\newcommand{\inputTikZfigure}[1]{\centering\includegraphics{figures/cache/#1}}

\fi

\DeclareMathOperator\huber{huber} 
\newcommand{\abs}[1]{\left\lvert#1\right\rvert}

\Crefname{equation}{Equation}{Equations}
\Crefname{figure}{Figure}{Figures}
\crefformat{equation}{Equation~#2#1#3}

\definecolor{orange}{rgb}{1.0,0.4,0}
\definecolor{skyblue}{rgb}{0.2,0.6,0.9}
\definecolor{placeholder}{rgb}{0.6,0.8,0.95}
\newcommand{\newnew}[1]{#1}
\newcommand{\NEWNEW}[1]{#1}
\newcommand{\NEW}[1]{#1}

\newcommand{\pose}{\mathbf{p}}

\newcommand{\I}{\mathcal{I}}

\newcommand{\s}{{\sigma}}
\newcommand{\cam}{\mathbf{o}}

\newcommand{\n}{\mathbf{n}}
\newcommand{\an}{\mathbf{a}}
\newcommand{\V}{\mathcal{V}}

\newcommand{\m}{\boldsymbol{\mu}}

\newcommand*\dif{\mathop{}\!\mathrm{d}} %

\newcommand{\dt}{\dif t}
\newcommand{\ds}{\dif s}

\title{EgoCap: Egocentric Marker-less Motion Capture with Two Fisheye Cameras}
\author{
	\begin{tabular}{cccc}
		Helge Rhodin$^\text{1}$ &
		Christian Richardt$^\text{1, 2, 3}$ &
		Dan Casas$^\text{1}$ &
		Eldar Insafutdinov$^\text{1}$ \\
		Mohammad Shafiei$^\text{1}$ &
		Hans-Peter Seidel$^\text{1}$ &
		Bernt Schiele$^\text{1}$ &
		Christian Theobalt$^\text{1}$\\[0.5em]
	\end{tabular}\\[0.5em]
	$^\text{1}$Max Planck Institute for Informatics\quad%
	$^\text{2}$Intel Visual Computing Institute\quad%
	$^\text{3}$University of Bath
}
\pdfauthor{Rhodin, Richardt, Casas, Insafutdinov, Shafiei, Seidel, Schiele and Theobalt}%


\setcopyright{acmlicensed}

\begin{document}

\teaser{
\includegraphics[width=\linewidth]{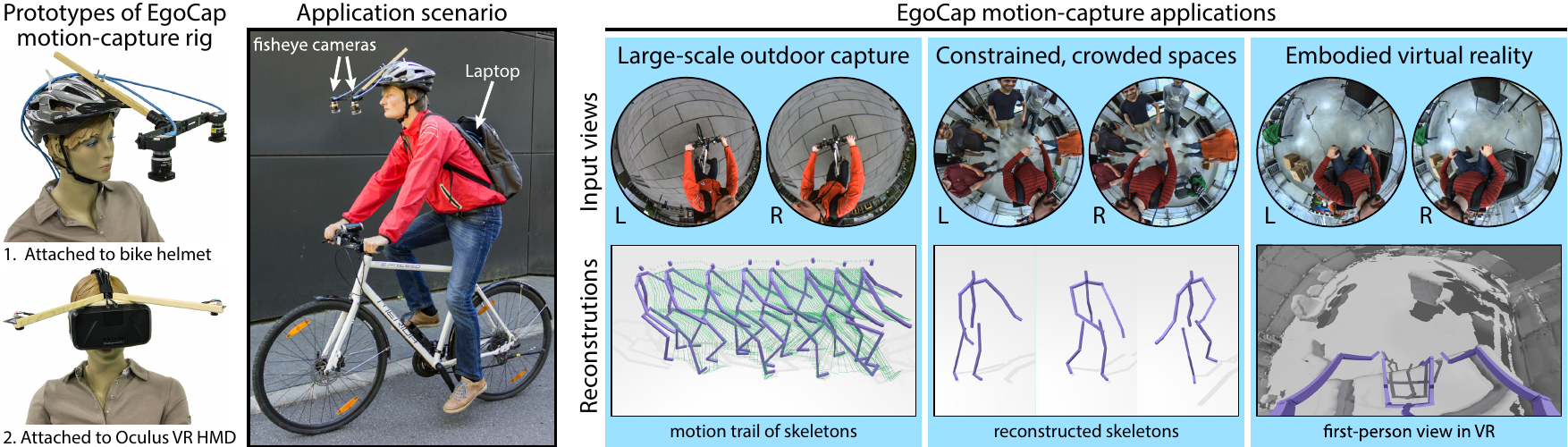}%
\caption{\label{fig:teaser}%
	We propose a marker-less optical motion-capture approach that only uses two head-mounted fisheye cameras (see rigs on the left).
	Our approach enables three new application scenarios:
	(1) capturing human motions in outdoor environments of virtually unlimited size,
	(2)~capturing motions in space-constrained environments, e.g. during social interactions, and
	(3) rendering the reconstruction of one's real body in virtual reality for embodied immersion.
}
}

\maketitle

\begin{abstract}

Marker-based and marker-less optical skeletal motion-capture methods use an \emph{outside-in} arrangement of cameras placed around a scene, with viewpoints converging on the center. 
They often create discomfort with marker suits, and their recording volume is \NEW{severely} restricted and often constrained to indoor scenes with controlled backgrounds. 
Alternative suit-based systems use several inertial measurement units or an exoskeleton 
to capture motion \NEWNEW{with an \emph{inside-in} setup, i.e. without external sensors}.
This makes capture independent of a confined volume, but requires substantial, often constraining, and hard to set up body instrumentation.
Therefore, we propose a new method for real-time, marker-less, and egocentric motion capture: \NEW{estimating the full-body skeleton pose} from a lightweight stereo pair of \NEW{fisheye} cameras attached to a helmet or virtual reality headset -- an \emph{optical inside-in} method, so to speak.
This allows full-body motion capture in general indoor and outdoor scenes, including crowded scenes with many people nearby,
which enables reconstruction in larger-scale activities.
Our approach combines the strength of a new generative pose estimation framework for fisheye views with a ConvNet-based body-part detector trained on a large new dataset. It is particularly useful in virtual reality to freely roam and interact, while seeing the fully motion-captured virtual body.

\end{abstract}


\keywordlist

\conceptlist

\vspace{1cm}
\section{Introduction}
\label{sec:intro}

Traditional optical skeletal motion-capture methods – both marker-based and marker-less – use several cameras typically placed around a scene in an \textit{outside-in} arrangement, with camera views approximately converging in the center of a confined recording volume.
This greatly constrains the spatial extent of motions that can be recorded; simply enlarging the recording volume by using more cameras, for instance to capture an athlete, is not scalable.
Outside-in arrangements also constrain the type of scene that can be recorded, even if it fits into a confined space.
If a recording location is too small, cameras can often not be placed sufficiently far away.
In other cases, a scene may be cluttered with objects or furniture, or other dynamic scene elements, such as people in close interaction, may obstruct a motion-captured person in the scene or create unwanted dynamics in the background.
In such cases, even state-of-the-art outside-in marker-less optical methods that succeed with just a few cameras, and are designed for less controlled and outdoor scenes \citep{ElhayAJTPABST2015}, quickly fail.
Scenes with dense social interaction were previously captured with outside-in camera
arrays of a few hundred sensors \citep{JooLTGNMKNS2015}, a very complex and difficult to scale setup.

These strong constraints on recording volume and scene density prevent the use of optical motion capture in the majority of real-world scenes.
This problem can partly be bypassed with \NEWNEW{\emph{inside-in}} motion-capture methods that use body-worn sensors \NEWNEW{exclusively} \cite{Menache:2010:UMC:1965363},
\NEW{such as the Xsens MVN inertial measurement unit suit.}
However, the special suit and cabling are obstructive and require tedious calibration.
\citet{ShiraPSSH2011} propose to wear 16 cameras placed on body parts facing \textit{inside-out}, and capture the skeletal motion through structure-from-motion relative to the environment.
\NEW{This clever solution requires instrumentation, calibration and a static background, but allows free roaming.
	This design was inspirational for our egocentric approach.}

\vspace{1cm}

\NEW{%
We propose EgoCap: an egocentric motion-capture approach that estimates full-body pose 
from a pair of optical cameras carried by lightweight headgear (see \cref{fig:teaser}).}
\NEWNEW{The body-worn cameras are oriented such that their field of view covers the user's body entirely, forming an arrangement that is independent of external sensors -- an \emph{optical inside-in} method, if you will.	
We show that our optical full-body approach overcomes many limitations of existing outside-in, inside-out and IMU-based inside-in methods.}
\NEW{It reduces the setup effort, enables free roaming, and minimizes body instrumentation.}
\NEW{EgoCap decouples the estimation of local body pose with respect to the headgear cameras and global headgear position, which we infer by \NEWNEW{inside-out} structure-from-motion on the scene background.} %

Our first contribution is a new \NEW{egocentric} inside-in sensor rig with only two head-mounted, downward-facing commodity video cameras with fisheye lenses (see \cref{fig:teaser}).
\newnew{While head-mounted cameras might pose a problem with respect to social acceptance and ergonomics in some scenarios, performances have not been hindered during our recordings and VR tests.}
The rig can be attached to a helmet or a head-mounted VR display, and, hence, requires less instrumentation and calibration
than other body-worn systems.
The stereo fisheye optics keep the whole body in view in all poses, despite the cameras' proximity to the body. %
We prefer conventional video cameras over IR-based RGB-D cameras, which were for example used for
egocentric hand tracking \citep{SridhMOT2015}, as
video cameras work indoors and outdoors, have lower energy consumption and are easily fitted
with the required fisheye optics.

Our second contribution is a new marker-less motion capture algorithm tailored to the strongly distorted egocentric fisheye views.
It combines a generative model-based skeletal pose estimation approach (\cref{sec:method}) with evidence from a trained ConvNet-based body part detector (\cref{sec:detections}).
\NEW{
The approach features an analytically differentiable objective energy that can be minimized efficiently,
is designed to work with unsegmented frames and general backgrounds, 
succeeds even on poses exhibiting notable self-occlusions (e.g. when walking), as the part detector predicts occluded parts, and enables recovery from tracking errors after severe occlusions.
}
Our third contribution is a new approach for automatically creating body part detection training datasets.
We record test subjects in front of green screen with an existing outside-in marker-less motion capture system to get ground-truth skeletal poses, which are reprojected into the simultaneously recorded head-mounted fisheye views to get 2D body part annotations.
We augment the training images by replacing the green screen with random background images, and vary the appearance in terms of color and shading by intrinsic recoloring \cite{MekaZRT2016}.
\NEW{With this technique, we annotate a total of 100,000 egocentric images of eight people in different clothing (\cref{sec:dataset}), with 75,000 images from six people used for training.
We publish the dataset for research purposes \cite{EgoCapData2016}.}

We designed and extensively tested two system prototypes featuring (1) cameras fitted to a bike helmet, and (2) small cameras attached to an Oculus Rift headset.
We show reliable egocentric motion capture, both off-line and in real time.
\NEW{The egocentric tracking meets the accuracy of outside-in approaches using 2--3 cameras; additional advances are necessary to match the accuracy of many-camera systems.}
\newnew{%
	In our egocentric setup, reconstructing the lower body is more challenging due to its larger distance and frequent occlusions, and is less accurate compared to the upper body in our experiments.}
\NEW{Nevertheless, we}
succeed in scenes that are challenging for \NEW{outside-in} approaches, such as close interaction with many people, as well outdoor and indoor scenes in cluttered environments with frequent occlusions, for example when working in a kitchen or at a desk.
We also show successful capturing in large volumes, for example of the skeletal motion of a cyclist.
The lightweight Oculus Rift gear is designed for egocentric motion capture for virtual reality, 
where the user can move in the real world to roam and interact in a virtual environment seen through a head-mounted display,
while perceiving increased immersion thanks to the rendering of the motion-captured body, which is not obtained with current HMD head pose tracking.
\section{Related Work}
\label{sec:related}

\paragraph{Suit-based Motion Capture}
Marker-based optical systems use a suit with passive retro-reflective spheres (e.g. Vicon) or active LEDs (e.g. PhaseSpace). 
Skeleton motion is reconstructed from observed marker positions in multiple cameras (usually 10 or more) in an outside-in arrangement,
producing highly accurate sparse motion data, even of soft tissue \citep{Park2008,loper2014mosh}, but \NEW{the external cameras} severely restrict the recording volume.
\NEW{For character animation purposes, where motions are restricted, use of motion sub-spaces can reduce requirements to six markers and two cameras \citep{chai2005performance}, or a single foot pressure-sensor pad \cite{yin2003footsee}, which greatly improves usability.
For hand tracking, a color glove and one camera \citep{wang2009real} is highly practical}.
Inertial measurement units (IMUs) fitted to a suit (e.g. Xsens MVN) \NEWNEW{allow free roaming and high reliability in cluttered scenes by inside-in motion capture, i.e. without requiring external sensors} \citep{TautgZKBWHMSE2011}.
\NEW{Combinations with ultrasonic distance sensors \cite{vlasic2007practical}, video input \citep{PonsMBHMSR2010,PonsMBGLMSR2011}, and pressure plates \cite{ha2011human} suppress the drift inherent to IMU measurements and reduce the number of required IMUs.}
\NEW{Besides drift, the instrumentation with IMU sensors is the largest drawback, causing long setup times and intrusion.}
Exoskeleton suits (e.g. METAmotion Gypsy) \NEW{avoid drift, but require} more cumbersome instrumentation.
Turning the standard outside-in capturing approach on its head, \citet{ShiraPSSH2011} attach 16 cameras to body segments in an inside-out configuration, and estimate skeletal motion from the position and orientation of each camera as computed with structure-from-motion.
\NEW{This clever solution -- which was inspirational for our egocentric approach -- allows free roaming although it requires instrumentation and a static background.}

\paragraph{Marker-less Motion Capture}
Recent years have seen great advances in marker-less optical motion-capture algorithms that track full-body skeletal motions, \NEW{reaching and outperforming the reconstruction quality of suit- and marker-based approaches} \citep{BreglM1998,TheobASST2010,MoeslHKS2011,HolteTTM2012}.
Marker-less approaches also typically use an outside-in camera setup, and were traditionally limited to controlled studio environments, or scenes with static, easy-to-segment background, using 8 or more cameras \citep[e.g.][]{urtasun2006temporal,Gall:2010,SigalBB2010,SigalIHB2012,StollHGST2011}.
Recent work is moving towards less controlled environments and outdoor scenes, also using fewer cameras \citep{AminARS2009,BurenSC2013,ElhayAJTPABST2015,RhodiRRST2015}, but still in an outside-in configuration.
\NEW{These approaches are well-suited for static studio setups, but} share the limitation of constrained recording volumes, and \NEW{reach} their limits in dense, crowded scenes.
\citet{JooLTGNMKNS2015} use a camera dome with 480 \NEW{outside-in} cameras for motion capture of closely interacting people, but domes do not scale to larger natural scenes.

\paragraph{Motion Capture with Depth Sensors}
\NEW{3D pose estimation is highly accurate and reliable when using multiple RGB-D cameras \cite{zhang2014leveraging}, and even feasible from a single RGB-D camera in real time} \citep[e.g.][]{Shotton:2011,Baak:2011,Wei:2012}.
However, many active IR-based depth cameras are unsuitable for outdoor capture, \NEW{have high energy consumption}, and equipping them with fisheye optics needed for our camera placement is hard.

\paragraph{Egocentric Motion Capture} %
In the past, egocentric \NEWNEW{inside-in} camera placements were used for tracking or model learning of certain parts of the body, for example of the face with a helmet-mounted camera or rig \citep{JonesFYMBIBD2011,WangCF2016}, of fingers from a wrist-worn camera \citep{Kim:2012}, 
or of eyes and eye gaze from cameras in a head-mounted rig \citep{Sugano:2015}.
\citet{Rogez2014} and \citet{SridhMOT2015} track articulated hand motion from body- or chest-worn RGB-D cameras.
Using a body-worn depth camera, \citet{YonemMOSST2015} extrapolate arm and torso poses from arm-only RGB-D footage.
\citet{JiangG2016} attempted full-body pose estimation from a chest-worn camera view by analyzing the scene, but without observing the user directly and at very restricted accuracy.
Articulated full-body motion capture with a lightweight head-mounted camera pair was not yet attempted.

\paragraph{First-person Vision} %
\NEW{A complementary research branch analyses the environment from first-person, i.e.~body-worn outward-facing cameras,
for activity recognition \citep[e.g.][]{FathiFR2011,kitani2011fast,OhnisKKH2016,ma2016going},
for learning engagement and saliency patterns of users when interacting with the real world \citep[e.g.][]{ParkJS2012,SuG2016},
and for understanding the utility of surrounding objects \cite{rhinehart2016learning}.
}
Articulated full-body tracking, or even only arm tracking, is not their goal, \NEW{but synergies of both fields appear promising}.

\paragraph{2D and 3D Pose Detection}

\NEW{Traditionally, 2D human pose estimation from monocular images is a two-stage process where coherent body pose is inferred from local image evidence \cite{YangPAMI2012,JohnsonCVPR2011}.
Convolutional networks (Conv\-Nets) brought a major leap in performance \cite{ChenY2014,JainTATB2014,JainTLB2015,TompsJLB2014,TosheS2014} and
recent models demonstrated that end-to-end prediction is possible
due to the large receptive fields capturing the complete pose context \citep{PishcITAAGS2016}.}
\citet{PfistCZ2015} and \citet{WeiRKS2016} allow for increased depth and learning of spatial dependencies between body parts
by layering multiple ConvNets.
We adopt the network architecture of \citet{InsafPAAS2016}, which builds on the recent success of residual networks \citep{HeZRS2016,NewelYD2016}, which further facilitate an increase in network depth.
\NEW{%
Recently, direct 3D pose estimation has emerged by lifting 2D poses to 3D \cite{YasinIKWG2016},
using mid-level posebit descriptors \cite{PonsMFR2014},
and motion compensation in videos \citep{Tekin:2016}, but estimates are still coarse.
}
Existing detection methods use simplified body models with few body parts to reduce the enormous cost of creating sufficiently large, annotated training datasets, do not generalize to new camera geometry and viewpoints, such as egocentric views, and results usually exhibit jitter over time.

\section{Egocentric Camera Design}
\label{sec:rig}

\begin{figure}
	\includegraphics[width=\linewidth]{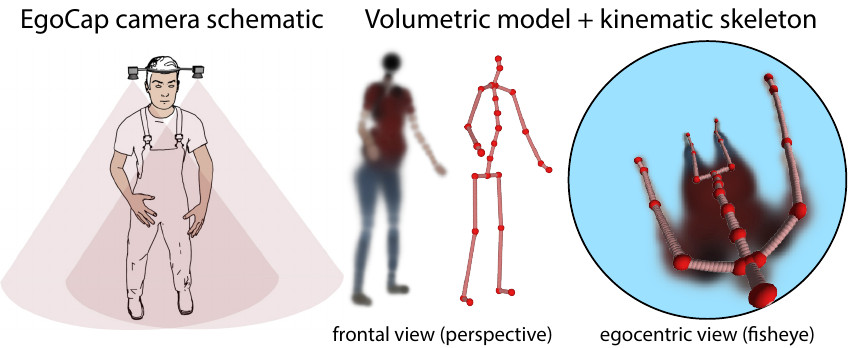}%
	\caption{\label{fig:volumetric-model}
		Schematic of EgoCap, our egocentric motion-capture rig (left), %
		visualization of the corresponding volumetric body model and kinematic skeleton (center),
		and the egocentric view of both in our head-mounted fisheye cameras (right).
	}%
\end{figure}

We designed a mobile egocentric camera setup to enable human motion capture within a virtually unlimited recording volume.
We attach two fisheye cameras rigidly to a helmet or VR headset, such that their field of view captures the user's full body, see \cref{fig:volumetric-model}.
The wide field of view allows to observe interactions in front and beside the user, irrespective of their global motion and head orientation, and without requiring additional sensors or suits.
The stereo setup ensures that most actions are observed by at least one camera, despite substantial self-occlusions of arms, torso and legs in such an egocentric setup.
A baseline of 30–40\,cm proved to be best in our experiments.
The impact of the headgear on the user's motion is limited as it is lightweight: our prototype camera rig for VR headsets (see \cref{fig:teaser}, bottom left) only adds about 65 grams of weight.
\section{Egocentric \NEWNEW{Full-Body} Motion Capture}
\label{sec:method}

\NEW{ 
Our egocentric setup separates human motion capture into two subproblems: (1) local skeleton pose estimation with respect to the camera rig, and (2) global rig pose estimation relative to the environment.
Global pose is estimated with existing structure-from-motion techniques, see \cref{sec:VR}.}
We formulate skeletal pose estimation as an analysis-by-synthesis-style optimization problem in the pose parameters $\pose^t$, 
that maximizes the alignment of a projected 3D human body model (\cref{sec:bodymodel}) with the human in the left $\I_\text{left}^t$ and the right $\I_\text{right}^t$ stereo fisheye views, at each video time step $t$.
We use a hybrid alignment energy combining evidence from a generative image-formation model, as well as from a discriminative detection approach. 
Our generative ray-casting-based image formation model is inspired by light transport in volumetric translucent media, and enables us 
to formulate a color-based alignment term in $\pose^t$ that is analytically differentiable and features an analytically differentiable formulation of 3D visibility (\cref{sec:iccv-model}). This model facilitates generative pose estimation with only two cameras, and we adapt it to the strongly distorted fisheye views. 
Our energy also employs constraints from one-shot joint-location predictions in the form of $E_\text{detection}$. 
These predictions are found with a new ConvNet-based 2D joint detector for head-mounted fisheye views, which is learned from a large corpus of annotated training data, and which generalizes to different users and cluttered scenes (\cref{sec:detections}).
The combined energy that we optimize takes the following form:
\begin{align}
E(\pose^t) \!=\! E_\text{color}(\pose^t) \!+\! E_\text{detection}(\pose^t) \!+\! E_\text{pose}(\pose^t) \!+\! E_\text{smooth}(\pose^t)
\text{.}
\label{eqn:objective}
\end{align}
Here, $E_\text{pose}(\pose^t)$ is a regularizer that penalizes violations of anatomical joint-angle limits as well as poses deviating strongly from the rest pose ($\pose \!=\! \mathbf{0}$):
\begin{align}
E_\text{pose}(\pose^t) = \ & \lambda_\text{limit} \!\cdot\! \Big(\! \max(0, \pose^t-\mathbf{l}_\text{upper})^2 + \max(0, \mathbf{l}_\text{lower}-\pose^t)^2 \Big) \nonumber \\
& + \lambda_\text{pose} \cdot \huber(\pose^t) \text{,}
\end{align} 
where $\mathbf{l}_\text{lower}$ and $\mathbf{l}_\text{upper}$ are \NEW{lower and upper} joint\NEW{-angle} limits, and
$\huber(x) \!=\! \sqrt{1 \!+\! x^2} \!-\! 1$ is the Pseudo-Huber loss function.
$E_\text{smooth}(\pose^t)$ is a temporal smoothness term:
\begin{align}
E_\text{smooth}(\pose^t) = \lambda_\text{smooth} \cdot \huber(\pose^{t-1} \!+\! \zeta(\pose^{t-1} \!-\! \pose^{t-2}) \!-\! \pose^t) \text{,} 
\end{align}
where $\zeta\!=$0.25 is a damping factor.
The total energy in \cref{eqn:objective} is optimized for every frame, as described in \cref{sec:optimization}.
In the following, we describe the generative and discriminative terms in more detail,
while omitting the temporal dependency $t$ in the notation for better readability.

We use weights $\lambda_\text{pose} \!=\! 10^{-4}$, $\lambda_\text{limit} \!=\! 0.1$ and $\lambda_\text{smooth} \!=\! 0.1$.

\subsection{Body Model}
\label{sec:bodymodel}

We model the 3D body shape and pose of humans in 3D using the approach proposed by \citet{RhodiRRST2015}, which represents the body volumetrically as a set of $N_q = 91$ isotropic Gaussian density functions distributed in 3D space.
Each Gaussian $G_q$ is parametrized by its standard deviation $\s_q$, location $\m_q$ in 3D space, density $c_q$ and color~$\an_q$, which define the Gaussian shape parameters.
The combined density field of the Gaussians, $\sum_q c_q G_q$, smoothly describes the volumetric occupancy of the human in 3D space, see \cref{fig:volumetric-model}.
Each Gaussian is rigidly attached to one of the bones of an articulated skeleton with 17 joints, whose pose is parameterized by 37 %
twist pose parameters \citep{Murray:1994}.

Shape and skeleton bone lengths need to be personalized to the tracked user prior to capturing. Commercial systems often use a dedicated initialization sequence at the start. 
Research papers on marker-less motion capture often treat initialization as a separate problem, and initialize models manually, which we could also do.
However, we propose a much more automated initialization procedure to reduce setup time and effort.
To this end, we adapt the approach of \citet{HelgeECCVSubmission}, who personalize a 3D parametric human shape model of Gaussian density and skeleton dimensions by fitting it to multi-view images using a volumetric contour alignment energy. 
We adapt this to our stereo fisheye setting.
In our egocentric setup 3--4 different user poses, \NEW{showing the bending of knees, elbows and wrists without any occlusion,} were sufficient for automatic shape and skeleton personalization, and only the automatically inferred Gaussian colors are manually corrected on body parts viewed at acute angles.

\subsection{Egocentric Volumetric Ray-Casting Model}
\label{sec:iccv-model}

For color-based model-to-image similarity, we use the ray-casting image formation model of the previously described volumetric body model \citep{RhodiRRST2015}.
We first describe image formation assuming a standard pinhole model, as in \citeauthor{RhodiRRST2015}, and then describe how we modify it for fisheye views.
A ray is cast from the camera center $\cam$ in direction $\n$ of an image pixel.
The visibility of a particular 3D Gaussian $G_q$ along the ray ($\cam + s\n$) is computed via
\begin{align}
\V_q(\cam,\n,\pose) \!=\! \!\int_{0}^{\infty} \!\!\!\!\!\!\!   \exp\!\left(\!\!-\!\!\int_0^s \! \sum_i \! G_i(\cam \!+\! t \n) \dt \! \right) \! G_q(\cam \!+\! s\n) \ds
\text{.}
\label{eqn:PreciseGaussianVisibility}
\end{align}
This formulation of visibility and color of a 3D Gaussian from the camera view is based on a model of light transport in heterogeneous translucent media \cite{Cerezo2005}. 
$\V_q$ is the fraction of light along the ray that is absorbed by Gaussian $G_q$.
We use this image-formation model in an energy term that computes the agreement of model and observation by summing the visibility-weighted color dissimilarity $d(\cdot,\cdot)$, which we describe in \cref{sec:appendix}, between image pixel color $\I(u,v)$ and the Gaussian's color $\an_q$:
\begin{equation}
E_\text{color}(\pose, \I) = %
\sum_{(u,v)} \! \sum_q d(\I(u,v), \an_q) \V_q(\cam,\n(u,v),\pose) 
\text{.}
\label{eqn:ColorTermSingle}
\end{equation}
Note that this formulation has several key advantages over previous generative models for image-based pose estimation.
It enables analytic derivatives of the pose energy, including a smooth analytically differentiable visibility model everywhere in pose space.
This makes it perform well with only a few camera views.
Previous methods often used fitting energies that are non-smooth or even lacking a closed-form formulation%
, requiring approximate recomputation of visibility (e.g. depth testing) inside an iterative optimization loop.
\citeauthor{RhodiRRST2015}'s formulation forms a good starting point for our egocentric tracking setting, as non-stationary backgrounds and occlusions are handled well.
However, it applies only to static cameras, does not support the distortion of fisheye lenses, and it does not run in real time.

\subsubsection{Egocentric Ray-Casting Model}
\label{sec:ego-model}

In our egocentric camera rig, the cameras move rigidly with the user's head.
In contrast to commonly used skeleton configurations, where the hip is taken as the root joint, our skeleton hierarchy is rooted at the head.
Like a puppet, the lower body parts are then relative to the head motion, see \cref{fig:volumetric-model}.
This formulation factors out the user's global motion, which can be estimated independently, see \cref{sec:VR}, and reduces the dimensionality of the pose estimation by 6 degrees of freedom.
By attaching the cameras to the skeleton root, the movable cameras are reduced to a static camera formulation such that \cref{eqn:PreciseGaussianVisibility} applies without modification.

Simply undistorting the fisheye images before optimization is impractical as resolution at the image center reduces and pinhole cameras cannot capture fields of view approaching 180 degrees \NEW{–} their image planes would need to be infinitely large.
To apply the ray-casting formulation described in the previous section to our egocentric motion-capture rig, with its 180\textdegree\ field of view, we replace the original pinhole camera model with the omnidirectional camera model of \citet{ScaraMS2006}.
The ray direction $\n(u,v)$ of a pixel $(u,v)$ is then given by
$\n(u,v) \!=\! [u, v, f(\rho)]^{\!\top}$, where $f$ is a polynomial of the distance $\rho$ of $(u, v)$ to the estimated image center.
We combine the energy terms for the two cameras (\cref{eqn:ColorTermSingle}) in our egocentric camera rig using
\begin{align}
E_\text{color}(\pose) =
	E_\text{color}(\pose, \I_\text{left}) + 
	E_\text{color}(\pose, \I_\text{right}) \text{.}
\label{eqn:ColorTerm}
\end{align}
These extensions also generalize the contour model of \citet{HelgeECCVSubmission} to enable egocentric body model initialization.
\subsection{Egocentric Body-Part Detection}
\label{sec:detections}

We combine the generative model-based alignment from the previous section with evidence from the discriminative joint-location detector of \citet{InsafPAAS2016}, trained on annotated egocentric fisheye images.
The discriminative component dramatically improves the quality and stability of reconstructed poses, \NEW{provides efficient recovery from tracking failures, and enables plausible tracking even under notable self-occlusions.
To apply \citeauthor{InsafPAAS2016}'s
body-part detector, which has shown state-of-the-art results on human pose estimation from outside-in RGB images, to the top-down perspective and fisheye distortion of our novel egocentric camera setup, the largest burden is to gather and annotate a training dataset that is sufficiently large and varied, containing tens of thousands of images.
As our camera rig is novel, there are no existing public datasets, and we therefore designed a method to automatically annotate real fisheye images by outside-in motion capture and to augment appearance with the help of intrinsic image decomposition.}
\subsubsection{Dataset Creation}
\label{sec:dataset}

\begin{figure}
	\includegraphics[width=\linewidth]{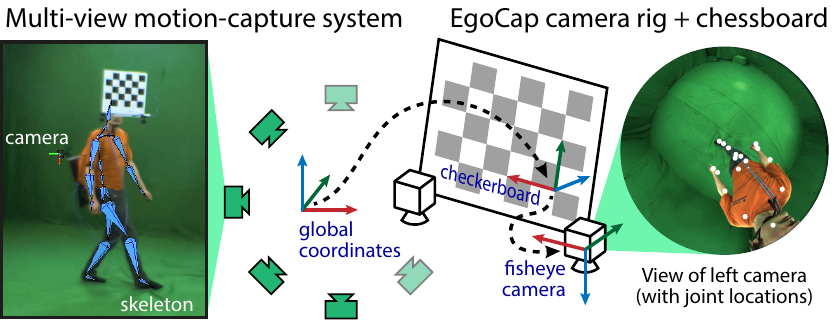}%
	\caption{\label{fig:joint-transfer}%
		\NEW{For database annotation,} the skeleton estimated from the multi-view motion capture system (left), is converted from global coordinates (center) into each fisheye camera's coordinate system (right) via the checkerboard.}%
\end{figure}

We propose a novel approach for semi-automatically creating large, realistic training datasets for body-part detection that comprise tens of thousands of camera images annotated with the joint locations of a kinematic skeleton and other body parts such as the hands and feet.
To avoid the tedious and error-prone manual annotation of locations in thousands of images, as in previous work, we use a state-of-the-art marker-less motion capture system (Captury Studio of The Captury)
to estimate the skeleton motion in 3D from eight stationary cameras placed around the scene.
We then project the skeleton joints into the fisheye images of our head-mounted camera rig.
\NEW{The projection} requires tracking the rigid motion of our head-mounted camera rig relative to the stationary cameras of the motion-capture system, for which we use a large checkerboard rigidly attached to our camera rig (\cref{fig:joint-transfer}).
We detect the checkerboard in all stationary cameras in which it is visible, and triangulate the 3D positions of its corners to estimate the pose and orientation of the camera rig.
Using \citeauthor{ScaraMS2006}'s camera distortion model, we then project the 3D joint locations into the fisheye images recorded by our camera rig.

\begin{figure}
	\inputTikZfigure{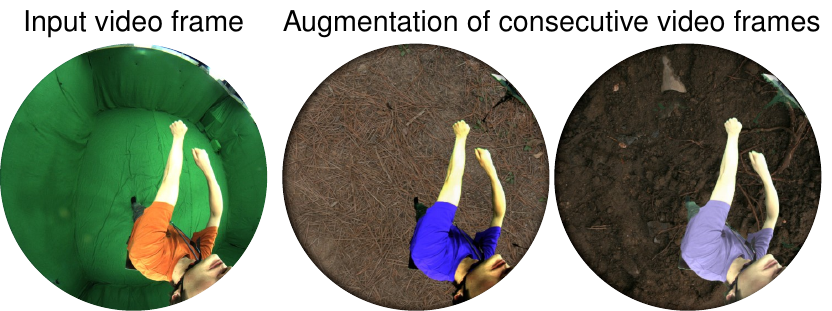}\vspace{-0.5em}%
	\caption{\label{fig:augmentation}%
		Illustration of our dataset augmentation using randomized backgrounds, intrinsic recoloring and gamma jittering.
		Note the varied shirt colors as well as brightness of the trousers and skin, which help prevent overtraining of the ConvNet-based joint detector.}%
\end{figure}

\paragraph{Dataset Augmentation}
We record video sequences of eight subjects performing various motions in a green-screen studio.
\NEW{%
For the training set, we replace the background of each video frame, using chroma keying, with a random, floor-related image from Flickr, as our fisheye cameras mostly see the ground below the tracked subject.
Please note that training with real backgrounds could give the CNN additional context, but is prone to overfitting to a (necessarily) small set of recorded real backgrounds.}
In addition, we augment the appearance of subjects by varying the colors of clothing, while preserving shading effects, using intrinsic recoloring \cite{MekaZRT2016}.
This is, to our knowledge, the first application of intrinsic recoloring for augmenting datasets.
We also apply a random gamma curve ($\gamma \!\in\! [\text{0.5}, \text{2}]$) to simulate changing lighting conditions.
We furthermore exploit the shared plane of symmetry of our camera rig and the human body to train a single detector on a dataset twice the size by mirroring the images and joint-location annotations of the right-hand camera to match those of the left-hand camera during training, and vice versa during motion capture. 
\NEW{Thanks to the augmentation, both background and clothing colors are different for every frame (see \cref{fig:augmentation}), which prevents overfitting to the limited variety of the captured appearances.
This results in a training set of six subjects and \texttildelow 75,000 annotated fisheye images.
Two additional subjects are captured and prepared for validation purposes.
}

\subsubsection{Detector Learning}
\label{sec:learning}

\NEW{%
Our starting point for learning an egocentric body-part detector for fisheye images is the 101-layer residual network \citep{HeZRS2016} trained by \citet{InsafPAAS2016} on the MPII Human Pose dataset \citep{AndriPGS2014}, which contains \texttildelow 19,000 internet images that were manually annotated in a crowd-sourced effort, and the Leeds Sports Extended dataset \citep{JohnsonCVPR2011} of 10,000 images.}
We remove the original prediction layers and replace them with ones that output 18 body-part heat maps\footnote{We jointly learn heat maps for the head and neck, plus the left and right shoulders, elbows, wrists, hands, hips, knees, ankles and feet.}.
The input video frames are scaled to a resolution of 640$\times$512 pixels, 
the predicted heat maps are of 8$\times$ coarser resolution.
We then fine-tune the ConvNet on our fisheye dataset 
for 220,000 iterations with a learning rate of 0.002, and drop it to 0.0002 for 20,000 additional iterations. 
The number of training iterations is chosen based on performance on the validation set. 
We randomly scale images during training by up to $\pm$15 to be more robust to variations in user size.
\Cref{fig:heatmapsOutput} (center) visualizes the computed heat maps for selected body parts.
\NEW{We demonstrate generalization capability to a large variety of backgrounds, changing illumination and clothing colors in \cref{sec:detections-eval}}.
\subsubsection{Body-Part Detection Energy}
\label{sec:detections-energy}

\begin{figure}
	\inputTikZfigure{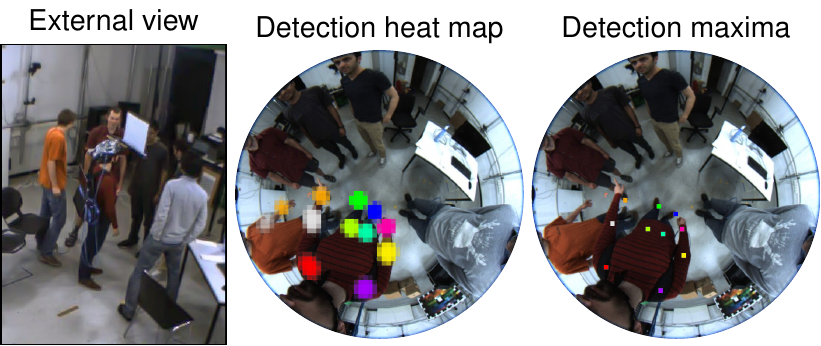}\vspace{-0.5em}%
	\caption{\label{fig:heatmapsOutput}%
		Color-coded joint-location detections on the \texttt{Crowded} sequence.
		For crowded scenes (left), detections can be multi-modal (center).
		However, the maximum (right) lies on the user.
		We exclude knee, hand and ankle locations for clearer visualization.
}%
\end{figure}

Inspired by \citet{ElhayAJTPABST2015}, who exploit detections in outside-in motion capture, we integrate the learned detections, in the form of heat maps as shown in \cref{fig:heatmapsOutput}, into the objective energy (\cref{eqn:objective}) as a soft constraint.
For each detection label, the location with maximum confidence, $(\hat{u},\hat{v})$, is selected and an associated 3D Gaussian is attached to the corresponding skeleton body part.
\NEW{%
This association can be thought of as giving a distinct color to each body-part label.
The Gaussian is used to compute the spatial agreement of the detection and body-part location in the same way as in the color similarity $E_\text{color}$, only the color distance $d(\cdot,\cdot)$ in \cref{eqn:ColorTermSingle} is replaced with the predicted detection confidence at $(\hat{u},\hat{v})$.
For instance, a light green Gaussian is placed at the right knee and is associated with the light green knee detection heat map at $(\hat{u},\hat{v})$, then their agreement is maximal when the Gaussian's center projects on $(\hat{u},\hat{v})$.
By this definition, $E_\text{detection}$ forms the sum over the detection agreements of all body parts and in both cameras.
We weight its influence by $\lambda_\text{detection} \!=\! 1/3$.}

\subsection{Real-Time Optimization}
\label{sec:optimization}

\citeauthor{RhodiRRST2015}'s volumetric ray-casting method \citeyearpar{RhodiRRST2015} models occlusion as a smooth phenomenon by integrating the visibility computations within the objective function instead of applying a depth test once before optimization.
While this is beneficial for optimizing disocclusions, it introduces dense pairwise dependencies between all Gaussians:
\NEW{
the visibility $\V_q$ (\cref{eqn:PreciseGaussianVisibility}) of a single Gaussian can be evaluated in linear time in terms of the number of Gaussians, $N_q$, but $E_\text{color}$ -- and its gradient with respect to all Gaussians -- has quadratic complexity in $N_q$.}

To nevertheless reach real-time performance, we introduce a new parallel stochastic optimization approach.
The ray-casting formulation allows a natural parallelization of $E_\text{detection}$ and $E_\text{color}$ terms and their gradient computation across pixels $(u, v)$ and Gaussians $G_q$.
We also introduce a traversal step, which determines the Gaussians that are close to each ray, and excludes distant Gaussians with negligible contribution to the energy.
These optimizations lead to significant run-time improvements, particularly when executed on a GPU,
but only enable interactive frame rates.

We achieve further reductions in run times by introducing a statistical optimization approach that is tailored to the ray-casting framework.
The input image pixels are statistically sampled for each gradient iteration step, as proposed by \citet{BlanzV1999}.
In addition, we sample the volumetric body model by excluding Gaussians from the 
\NEW{gradient computation at random, individually for each pixel}, which improves the optimization time to 10 fps and more.

\section{Evaluation}
\label{sec:evaluation}

\subsection{Hardware Prototypes}
\label{sec:prototype}

We show the two EgoCap prototypes used in this work in \cref{fig:teaser} (left).
\emph{EgoRig1} consists of two fisheye cameras attached to a standard bike helmet.
It is robust and well-suited for capturing outdoor activities and sports.
\emph{EgoRig2} builds on a lightweight wooden rig that holds two consumer cameras and is glued to an Oculus VR headset.
It weighs only 65\,grams and adds minimal discomfort on the user.
Both prototypes are equipped with 180\textdegree\ fisheye lenses and record with a resolution of 1280$\times$1024\,pixels at 30\,Hz.
 Note that the checkerboard attached to \emph{EgoRig1} in several images is not used for tracking (only used in training and validation dataset recordings). 

\begin{figure}
	\inputTikZfigure{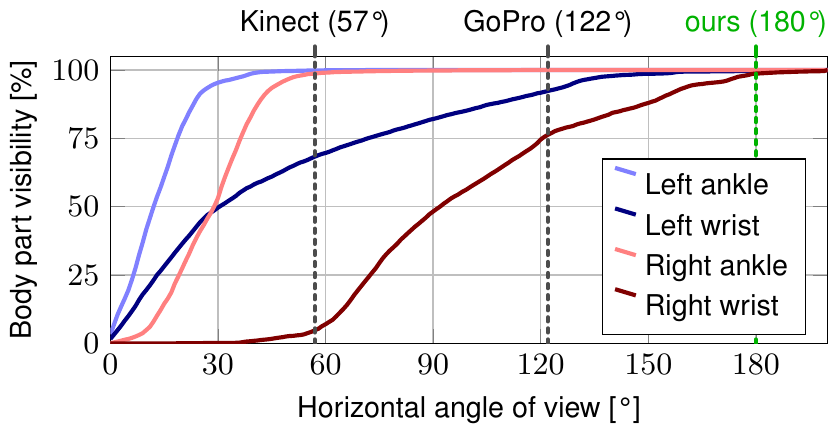}\vspace{-0.5em}%
	\caption{\label{fig:fov}%
		Visibility of selected body parts for different camera angles of view, for the left-hand camera in our rig over a 5-minute recording.
		Seeing the right wrist 95 percent of the time requires an angle of view in excess of 160\textdegree, which is only practical with fisheye lenses.}
\end{figure}

\paragraph{Body-Part Visibility}
For egocentric tracking of unconstrained motions, the full 180\textdegree\ field of view is essential for egocentric tracking.
We evaluate the visibility of selected body parts from our egocentric rig with different (virtual) field-of-view angles in \cref{fig:fov}.
Only at 180 degrees are almost all body parts captured, otherwise even small motions of the head can cause the hand to leave the recording volume.
The limited field of view of existing active depth sensors of 60–80 degrees restricts their applicability to egocentric motion capture in addition to their higher energy consumption and interference with other light sources.

\subsection{Runtime}
\label{sec:runtime}

For most tracking results, we use a resolution of 128$\times$128\,pixels and 200 gradient-descent iterations.
Our CPU implementation runs at ten seconds per frame on a Xeon E5-1620 3.6\,GHz, which is similar to run times reported by \citet{RhodiRRST2015}.
Straightforward parallelization on the GPU reduces run times to two seconds per frame.
The body-part detector runs on a separate machine, and processes 6 images per second on an Nvidia Titan GPU and a Xeon E5-2643 3.30\,GHz.

For some experiments (see \cref{sec:VR}), we use a resolution of 120$\times$100\,pixels and enable stochastic optimization.
Then, purely color-based optimization reaches 10 to 15\,fps for 50 gradient iterations (2–3\,ms per iteration), i.e. close to real-time performance.
Our body-part detector is not optimized for speed and cannot yet run at this frame rate, but its implementation could be optimized for real-time processing, so a real-time end-to-end approach would be feasible without algorithmic changes.

\subsection{Body-Part Detections}
\label{sec:detections-eval}

We first evaluate the learned body-part detectors, irrespective of generative components, using the percentage of correct keypoints (PCK) metric \citep{Sapp2013,TompsJLB2014}.
We evaluate on a validation set, \texttt{Validation2D}, of 1000 images from a 30,000-frame sequence of two subjects that are not part of the training set and wear dissimilar clothing.
\texttt{Validation2D} is augmented with random backgrounds using the same procedure as for the training set, such that the difficulty of the detection task matches the real-world sequences. \NEW{We further validated that overfitting to augmentation is minimal, by testing on green-screen background, with equivalent results.}
\tabcolsep 1.5pt
\begin{table}[tbp]
\caption[]{\label{tab:pose2d}%
	Part detection accuracy in terms of the percentage of correct keypoints (PCK) on the validation dataset \texttt{Validation2D} of 1000 images, evaluated at 20\,pixel threshold for three ConvNets trained with different data augmentation strategies (\cref{sec:dataset}).
	AUC is area under curve evaluated for all thresholds up to 20\,pixels.
}\vspace{-0.25cm}
 \scriptsize
  \centering
  \begin{tabular}{@{} l c ccc ccc |c|c@{}}
  	\toprule
  	Training dataset setting           & Head          & Sho.          &     Elb.      &     Wri.      & Hip           &      Knee      &      Ank.      & PCK           &      AUC      \\ \midrule
  	green-screen background            & 75.5          & 46.8          &     18.8      &     13.6      & 17.4          & \phantom{0}7.2 & \phantom{0}4.5 & 22.4          &     10.0      \\
  	\quad + background augmentation    & 84.7          & 87.5          &     90.9      &     89.1      & 97.7          &      94.2      &      86.4      & 89.5          &     56.9      \\
  	\quad \quad + intrinsic recoloring & \textbf{86.2} & \textbf{96.1} & \textbf{93.6} & \textbf{90.1} & \textbf{99.1} & \textbf{95.8}  & \textbf{90.9}  & \textbf{92.5} & \textbf{59.4} \\ \bottomrule
  \end{tabular}

\end{table}

\begin{figure}
	\centering
	\begin{tabular}{@{\hspace{0pt}}c@{\hspace{6pt}}c@{\hspace{0pt}}}  
		\includegraphics[width=0.48\linewidth]{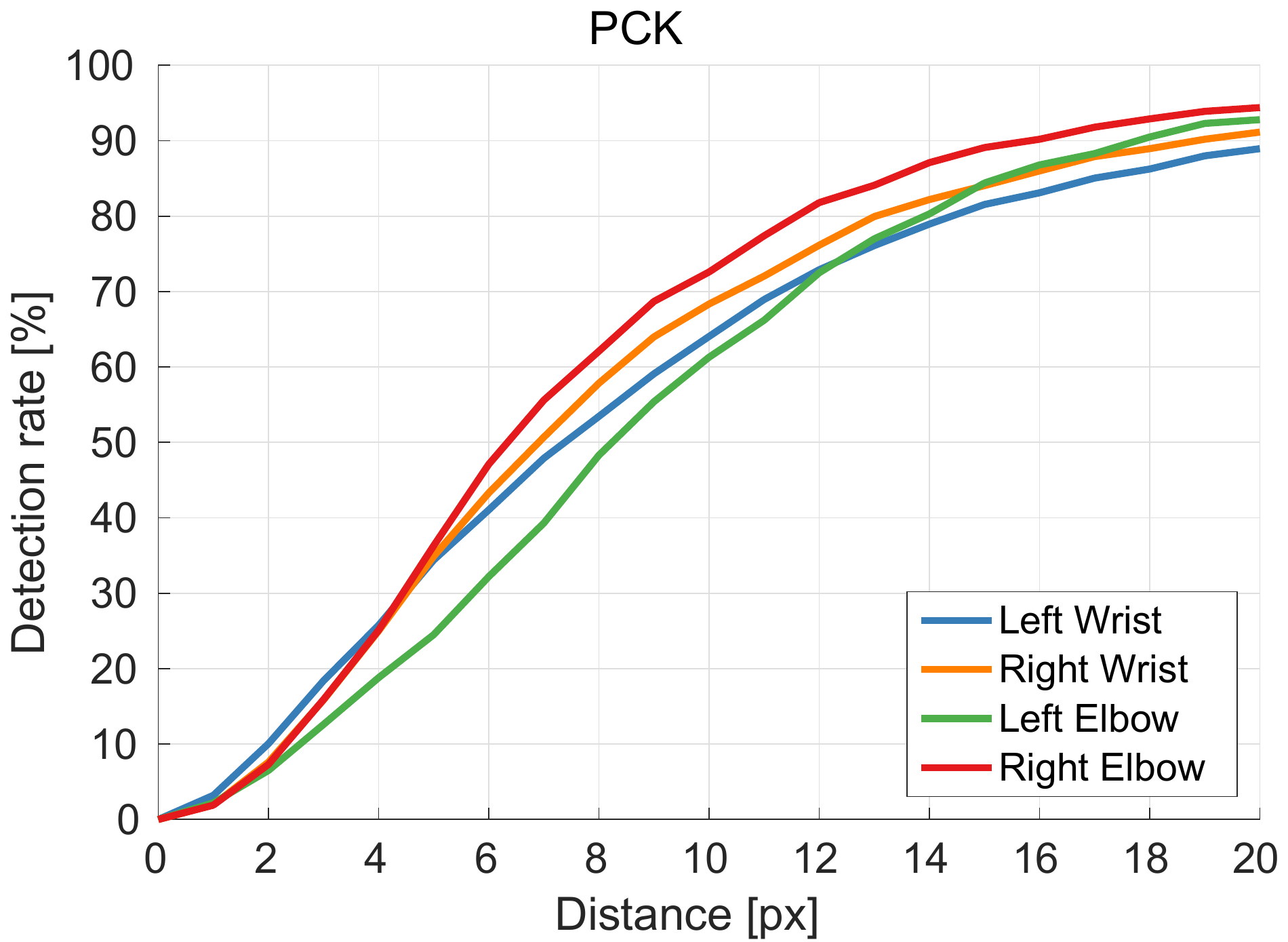}&
		\includegraphics[width=0.48\linewidth]{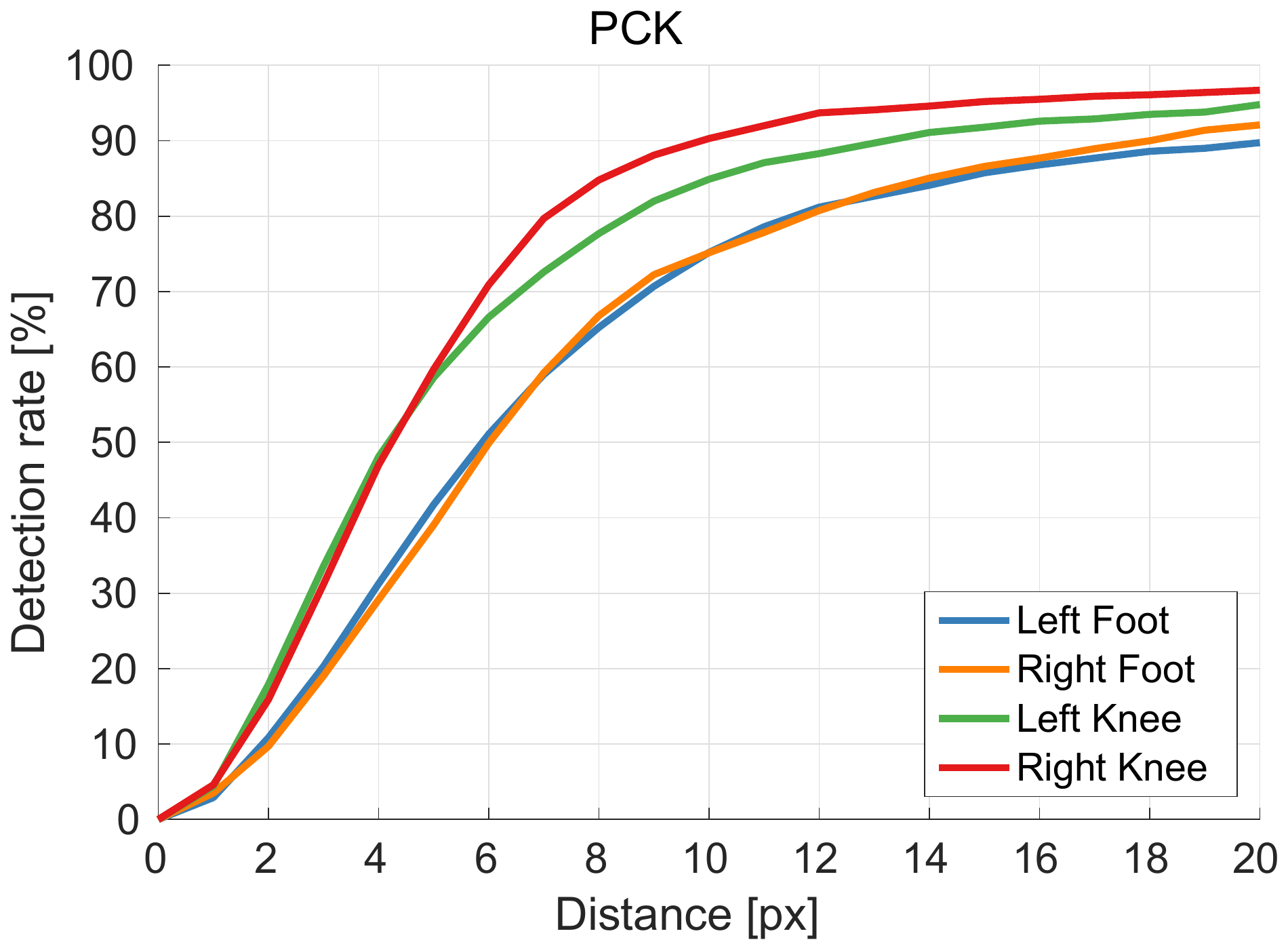}\\
		(a) Arm joints & (b) Leg joints \\
	\end{tabular}
	\vspace{-0.1em}
	\caption{Pose estimation results in terms of percentage of correct keypoints (PCK) for different distance thresholds on \texttt{Validation2D}.}
	\label{fig:pck-curves}
\end{figure}

\begin{figure*}
	\includegraphics[width=\linewidth]{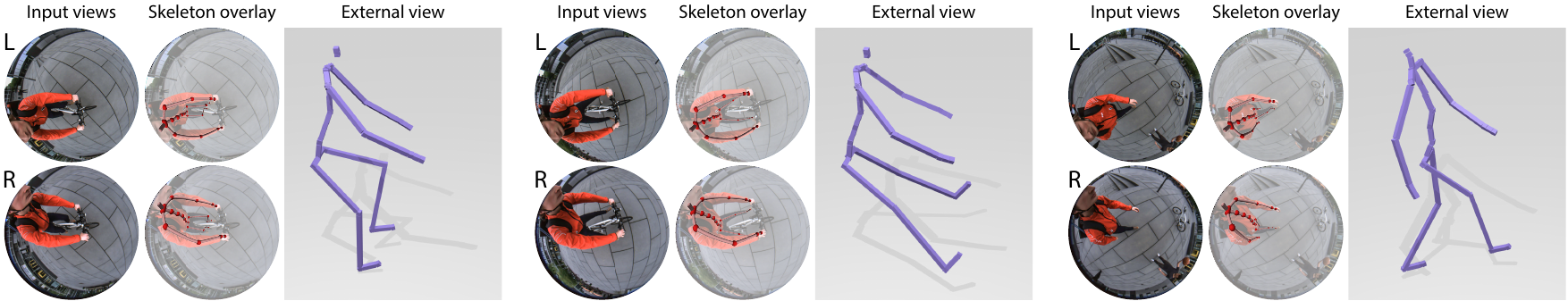}%
	\caption{\label{fig:outdoor-results}%
		EgoCap enables outdoor motion capture with virtually unconstrained extent.
		Full-body pose is accurately estimated for fast \texttt{Biking} (left and center) and for unconstrained \texttt{Walk} (right).
		The model is tailored to handle the present occlusions and strong image distortion.
	}\vspace{-1em}
\end{figure*}

\paragraph{Dataset Augmentations}
\Cref{tab:pose2d} presents the evaluation of proposed data augmentation strategies.
Background augmentation during training brings a clear improvement.
It provides a variety of challenging negative samples for the training of the detector, which is of high importance.
Secondly, the performance is further boosted by employing intrinsic video for cloth recoloring, which additionally increases the diversity of training samples.
The improvement of about two percent is consistent across all body parts.

\paragraph{Detection Accuracy}
\Cref{fig:pck-curves} contains the plots of PCK at different distance thresholds for arms and legs evaluated on sequence \texttt{Validation2D}.
We achieve high accuracy, with slightly lower detection reliability of terminal limbs (wrists, feet).
This can either be due to more articulation or, in case of the feet, due to higher occlusion by knees and their small appearance due to the strong fisheye distortion.
\NEW{%
The 2D detection accuracy of feet and wrists is comparable, even though feet are further away, and similar pixel error hence translates to larger 3D errors, as evaluated in the next section.}
\NEW{%
We additionally evaluated the training set size.
We found that subject variation is important: using only three out of six subjects, the PCK performance dropped by 2.5 percent points.
Moreover, using a random subset of 10 of the original database size reduces the PCK by 2 points, i.e. using more than three frames per second is beneficial.
Using a 50 subset did not degrade performance, showing that consecutive frames are not crucial for our per-frame model, but could be beneficial for future research, such as for temporal models.}

\subsection{3D Body Pose Accuracy}
\label{sec:comparisons}

Our main objective is to infer 3D human pose from the egocentric views, despite occlusions and strong fisheye image distortions.
\NEW{We quantitatively evaluate the 3D body pose accuracy of our approach on two sequences, \texttt{ValidationWalk} and \texttt{ValidationGest}.} Ground-truth data is obtained with the Captury Studio, a state-of-the-art marker-less commercial multi-view solution with eight video cameras \NEW{and 1--2\,cm accuracy}.
The two systems are used simultaneously and their relative transformation is estimated with a reference checkerboard, see \cref{fig:joint-transfer}.
We experimented with raw green-screen and with randomly replaced background.
Error values are estimated as the average Euclidean 3D distance over \NEW{17 joints, including all joints with detection labels, except the head}.
Reconstructions on green and replaced backgrounds are both \NEW{7$\pm$1\,cm} for a challenging 250-frame walking sequence with occlusions, and \NEW{7$\pm$1\,cm} on a long sequence of \NEW{750} frames of gesturing and interaction.
\NEW{%
During gesturing, where arms are close to the camera, upper body (shoulder, elbow, wrist, finger) joint accuracy is higher than for the lower body (hip, knee, ankle, and toe) with 6\,cm and 8\,cm average error, respectively.
During walking, upper and lower body error is similar with 7\,cm.
}
Please note that slight differences in skeleton topology between ground truth and EgoCap exist, which might bias the errors.

Despite the difficult viewing angle and image distortion of our egocentric setup, the overall 3D reconstruction error is comparable to state-of-the-art results of outside-in approaches  \cite{RhodiRRST2015,ElhayAJTPABST2015,AminARS2009,SigalBB2010,belagiannis20143d}, which reach 5–7\,cm accuracy from two or more cameras, but only in small and open recording volumes, and for static cameras.
In contrast, our algorithm scales to very narrow and cluttered scenes (see \cref{fig:crowded-results}) as well as to wide unconstrained performances (see \cref{fig:outdoor-results}).
No existing algorithm is directly applicable to these conditions and the strong distortions of the fisheye cameras, precluding a direct comparison.
Closest to our approach is the fundamentally off-line inside-out method of \citet{ShiraPSSH2011}, who use 16 body-worn cameras facing outwards, reporting a mean joint position error of 2\,cm on a slowly performed indoor walking sequence.
Visually, their outdoor results show similar quality to our reconstructions, although we require fewer cameras, and can handle crowded scenes.
\NEW{It depends on the application whether head gear or body-worn cameras less impair the user's performance.}

\subsection{Model Components}

Our objective energy consists of detection, color, smoothness, and pose prior terms.
Disabling the smoothness term increases the reconstruction error on the validation sequences by 3\,cm.
Without the color term, accuracy is reduced by 0.5\,cm.
We demonstrate in the supplemental video that the influence of the color term is more significant in the outdoor sequences for motions that are very dissimilar to the training set.
Disabling the detection term removes the ability to recover from tracking failures, which are usually unavoidable for fully automatic motion capture of long sequences with challenging motions.
High-frequency noise is filtered with a Gaussian low-pass filter of window size 5.

\section{Applications}
\label{sec:applications}

\NEW{We further evaluate our approach in three application scenarios with seven sequences of lengths of up to 1500 frames using \emph{EgoRig1}, 
in addition to the three quantitative evaluation sequences.}
\NEW{The captured users wear clothes not present in the training set.}
The qualitative results are best observed in the supplemental video.

\begin{figure*}
	\includegraphics[width=\linewidth]{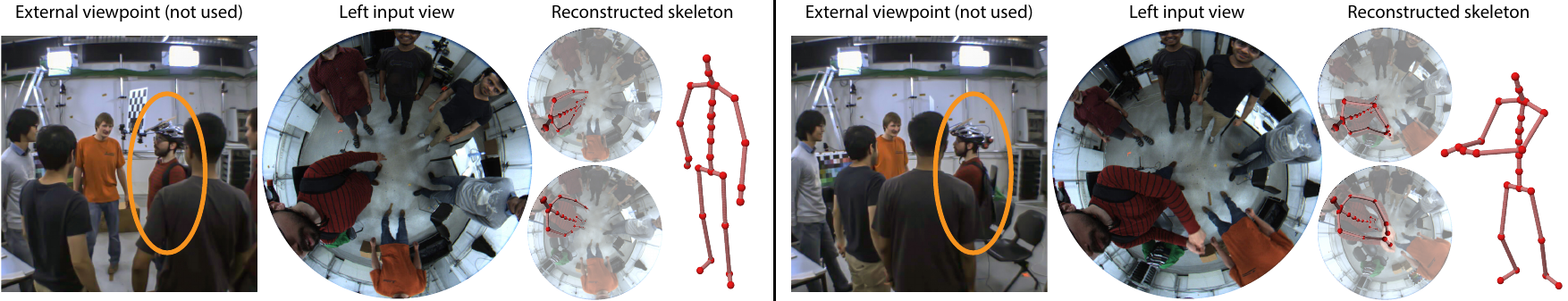}%
	\caption{\label{fig:crowded-results}%
		Capturing social interaction in crowded scenes is of importance, but occlusions pose difficulties for existing outside-in approaches (left). The egocentric view enables 3D pose estimation, as demonstrated on the \texttt{Crowded} sequence. The visible checkerboard is not used.
	}\vspace{-1em}
\end{figure*}

\subsection{Unconstrained/Large-Scale Motion Capture}

We captured a \texttt{Basketball} sequence outdoors, which shows quick motions, large steps on a steep staircase, and close interaction of arms, legs and the basketball (supplemental video).
We also recorded an outdoor \texttt{Walk} sequence with frequent arm-leg self-occlusions (\cref{fig:outdoor-results}, right).
With EgoCap, a user can even motion capture themselves while riding a bike in a larger volume of space (\texttt{Bike} sequence, \cref{fig:outdoor-results}, left and center).
The pedaling motion of the legs is nicely captured, despite frequent self-occlusions; the steering motion of the arms and the torso is also reconstructed.
Even for very fast absolute motions, like this one on a bike, our egocentric rig with cameras attached to the body leads to little motion blur, which challenges outside-in optical systems.
All this would have been \NEW{difficult} with \NEW{alternative} motion-capture approaches.

Note that our outdoor sequences also show the resilience of our method to different appearance and lighting conditions, as well as the generalization of our detector to a large range of scenes.

\begin{figure}
	\includegraphics[width=\linewidth]{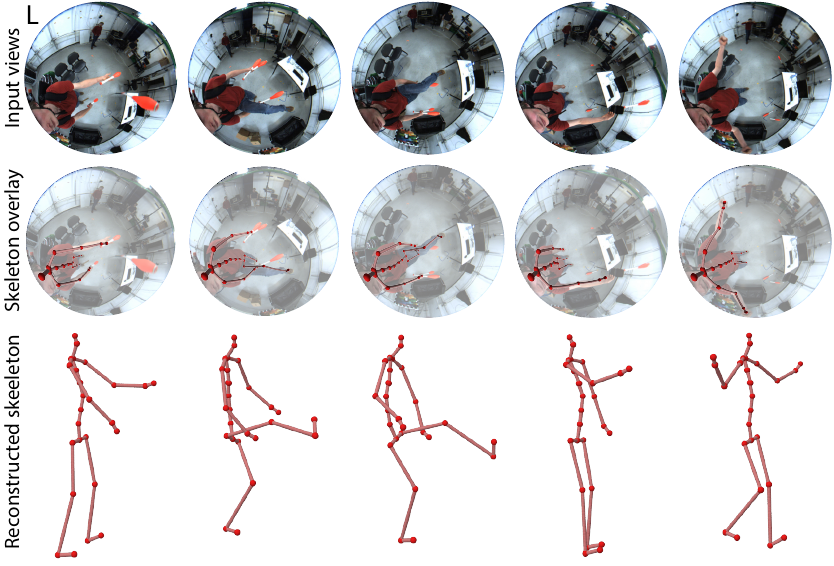}%
	\caption{\label{fig:juggling-results}%
		Reconstruction results on the \texttt{Juggler} sequence, showing one input view and the estimated skeleton.
		Despite frequent self-occlusions, our approach robustly recovers the skeleton motion.
	}\vspace{-1em}
\end{figure}

\subsection{Constrained/Crowded Spaces}

We also tested EgoCap with \emph{EgoRig1} for motion capture on the \texttt{Crowded} sequence, where many spectators are interacting and occluding the tracked user from the outside (\cref{fig:crowded-results}).
In such a setting, as well as in settings with many obstacles and narrow sections, outside-in motion capture, even with a  dense camera system, would be difficult.
In contrast, EgoCap captures the skeletal motion of the user in the center with only two head-mounted cameras. 

The egocentric camera placement is well-suited for capturing human-object interactions too, such as the juggling performance \texttt{Juggler} (\cref{fig:juggling-results}). Fast throwing motions as well as occlusions are handled well.
The central camera placement ensures that objects that are manipulated by the user are always in view.

\subsection{Tracking for Immersive VR}\label{sec:VR}

We also performed an experiment to show how EgoCap could be used in immersive virtual reality (VR) applications. 
To this end, we use \emph{EgoRig2} attached to an Oculus VR headset and track the motion of a user wearing it. 
We build a real-time demo application running at up to 15\,fps, showing that real-time performance is feasible with additional improvements on currently unoptimized code.
In this \texttt{Live} test, we only use color-based tracking \NEW{of the upper body}, without detections, as the detector code is not yet optimized for speed.
The \texttt{Live} sequence shows that body motions are tracked well, and that with such an even more lightweight capture rig, geared for HMD-based VR, egocentric motion capture is feasible.
In the supplemental video, we show an additional application sequence `\texttt{VR}', in which the the user can look down at their virtual self while sitting down on a virtual sofa.
Current HMD-based systems only track the pose of the display; our approach adds motion capture of the wearer's full body, which enables a much higher level of immersion.

\begin{figure}
	\includegraphics[trim=0cm 2.5cm 3cm 0cm,clip,width=\linewidth]{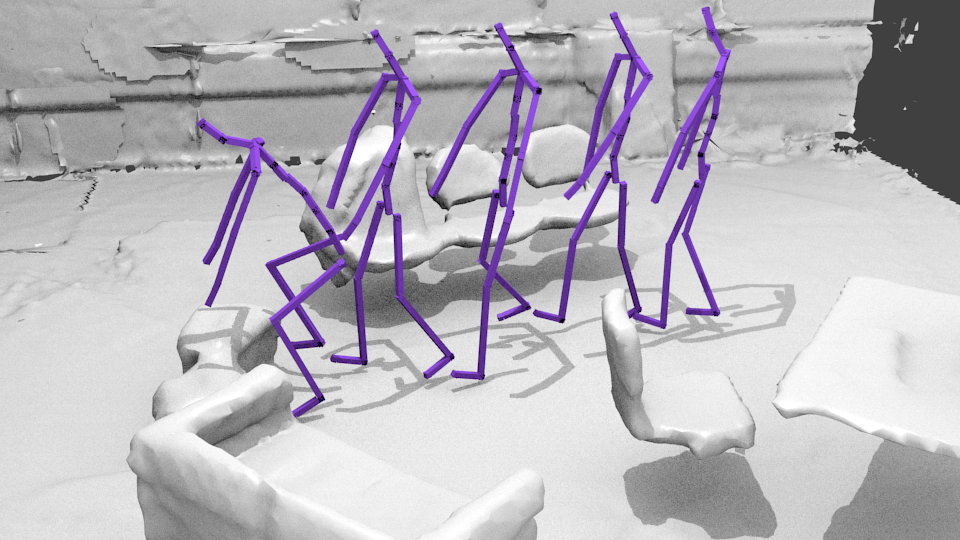}
	\caption{%
		\NEW{Complete motion-capture example \texttt{VR}, in which our egocentric pose tracking is combined with global pose tracking using structure-from-motion, shown as a motion sequence in a 3D reconstruction of the scene. In a VR scenario, this would allow free roaming and interaction with virtual objects.}
	}\vspace{-1em}
	\label{fig:global-tracking}
\end{figure}

\paragraph{\NEW{Global Pose Estimation}}
\NEW{For free roaming, the global rig pose can be tracked independently of external devices using structure-from-motion in the fisheye views. 
We demonstrate combined local and global pose estimation on the \texttt{Biking}, \texttt{Walk}, and \texttt{VR} sequence, using the structure-from-motion implementation of \citet{moulon2013global} provided in the OpenMVG library, see \cref{fig:global-tracking} and the accompanying video.
Such complete motion capture paves the way for immersive roaming in a fully virtual 3D environment.}

\section{Discussion \NEW{and Limitations}}
\label{sec:discussion}
We developed the first stereo egocentric motion-capture approach for indoor and outdoor scenes, that also works well for very crowded scenes.
The combination of generative and detection-based pose estimation make it fare well even under poses with notable self-occlusions. 
Similar to other outside-in optical methods, tracking under occlusions by objects in the environment, such as a table, may lead to tracking failures.
However, the detections enable our tracker to quickly recover from such occlusion failures.
Interestingly, the egocentric fisheye camera setup provides stronger perspective cues for motion towards and away from the camera than with normal optics. 
The perspective effect of the same motion increases with proximity to the camera. 
For instance, bending an arm is a subtle motion when observed from an external camera, 
but when observed in proximity, the same absolute motion causes large relative motion, manifesting in large displacements and scaling of the object in motion. 

The algorithm in this paper focuses on an entirely new way of capturing the full egocentric skeletal body pose, \NEW{that is decoupled from} global pose and rotation relative to the environment. 
\NEW{Global pose can be inferred separately by structure-from-motion from the fisheye cameras or is provided by HMD tracking in VR applications.} 
Fisheye cameras keep the whole body in view, but cause distortions reducing the image resolution of distant body parts such as the legs. 
Therefore, tracking accuracy of the upper body \NEW{is slightly} higher than that of the lower body. 
Also, while overall tracking accuracy of our research prototype is still lower than with commercial outside-in methods, it shows a new path towards 
more unconstrained capture in the future.
Currently, we have no real-time end-to-end prototype. We are confident that this would be feasible without algorithm redesign, yet felt \NEW{that real-time performance} is not essential to 
demonstrate the algorithm and its general feasibility. 

Our current prototype systems may still be a bit bulky, but much stronger miniaturization \NEW{becomes} feasible in mass production;
the design of \emph{EgoRig2} shows this possibility.
\NEW{Some camera extension is required for lower-body tracking and might pose a problem with respect to social acceptance and \newnew{ergonomics} for some applications;
However, we did not encounter practical issues during our recordings and VR tests, as users naturally keep the area in front of their head clear to not impair their vision.
Moreover, handling changing illumination is still an open problem for motion capture in general and is not the focus of our work.
For dynamic illumination, the color model would need to be extended. However, the CNN performs one-shot estimation and does not suffer from illumination changes.
The training data also contains shadowing from the studio illumination, although extreme directional light might still cause inaccuracies.
Additionally, loose clothing, such as a skirt, is not part of the training dataset and hence likely to reduce pose accuracy.
}

\section{Conclusion}
\label{sec:conclusion}

We presented EgoCap, the first approach for marker-less egocentric full-body motion capture with a head-mounted fisheye stereo rig. 
It is based on a pose optimization approach that jointly employs two components. 
The first is a new generative pose estimation approach 
based on a ray-casting image formation model enabling an analytically differentiable alignment energy and visibility model. 
The second component is a new ConvNet-based body-part detector for fisheye cameras that was trained on the first automatically annotated real-image training dataset of egocentric fisheye body poses. 
EgoCap's lightweight on-body capture strategy bears many advantages over other motion-capture methods. 
It enables motion capture of dense and crowded scenes, and reconstruction of large-scale activities that would not fit into the constrained recording volumes of outside-in motion-capture methods.
It requires far less instrumentation than suit-based or exoskeleton-based approaches. 
EgoCap is particularly suited for HMD-based VR applications; two cameras attached to an HMD enable full-body pose reconstruction of your own virtual body \NEW{to pave the way for immersive VR experiences and interactions}.

\section*{Acknowledgements}

We thank all reviewers for their valuable feedback, Dushyant Mehta, 
James Tompkin, and The Foundry for license support.
This research was funded by the ERC Starting Grant project CapReal (335545).

\bibliographystyle{acmsiggraphnat}
\bibliography{EgoTrack}

\appendix\section{Implementation Details}
\label{sec:appendix}
\label{sec:implementation}

\paragraph{Color Dissimilarity}
For measuring the dissimilarity $d(\mathbf{m}, \mathbf{i})$ of model color $\mathbf{m}$ and image pixel color $\mathbf{i}$ in \cref{eqn:ColorTermSingle}, we use the HSV color space (with all dimensions normalized to unit range) and combine three dissimilarity components:
\begin{enumerate}
\item
For saturated colors, the color dissimilarity $d_s$ is computed using the squared (minimum angular) hue distance. Using the hue channel alone gains invariance to illumination changes.

\item
For dark colors, the color dissimilarity $d_d$ is computed as twice the squared value difference, i.e. $d_d(\mathbf{m}, \mathbf{i}) \!=\! 2(m_v \!-\! i_v)^2$. Hue and saturation are ignored as they are unreliable for dark colors.

\item
For gray colors, the distance $d_g$ is computed as the sum of absolute value and saturation difference, i.e. $d_g(\mathbf{m}, \mathbf{i}) \!=\! \abs{m_v \!-\! i_v} \!+\! \abs{m_s \!-\! i_s}$. Hue is unreliable and thus ignored.
\end{enumerate}
We weight these three dissimilarity components by $w_s \!=\! \sqrt{m_s} / Z$, $w_d \!=\! \max(0, 0.5 \!-\! m_v) / Z$ and $w_g \!=\! \max(0, 0.5 \!-\! m_s) / Z$ respectively, where $Z$ normalizes the sum of these weights to unity.
The total dissimilarity is computed by $d(\mathbf{m}, \mathbf{i}) \!=\! \phi(w_s d_s \!+\! w_d d_d \!+\! w_g w_g)$ where $\phi(x) \!=\! 1 \!-\! (1 \!-\! x)^4 (8x \!+\! 2)$ is a smooth step function.
We employ a two-sided energy, i.e. $E_\text{color}$ can be negative:
For dissimilar colors, $d \approx 1$ and approaches $-1$ for similar colors.

\end{document}